\documentclass[11pt,a4paper]{article}
\usepackage[hyperref]{naaclhlt2019}
\usepackage{times}
\usepackage{latexsym}
\usepackage{url}
\usepackage{microtype}
\usepackage{scalefnt}
\usepackage{xspace}
\newcommand{\eg}{\textit{e.g.}\xspace}
\newcommand{\ie}{\textit{i.e.}\xspace}

\newcommand{\pnull}{\phantom{0}}
\newcommand{\cf}{\textit{cf.}\xspace}
\newcommand{\F}{$\text{F}_1$\xspace}

\newcommand{\tw}[2]{%
  \setlength{\fboxsep}{0pt}%
  \colorbox{red!#2!white}{#1}%
}
\usepackage{paralist}
\usepackage{multirow,graphicx}
\usepackage{booktabs} 
\usepackage{tabularx}
\usepackage{amsmath}
\usepackage{amsfonts}
\aclfinalcopy
\def\aclpaperid{334}

\title{Adversarial Training for Satire Detection:\\ Controlling for Confounding Variables}

\author{Robert McHardy, Heike Adel \and Roman Klinger \\
 Institut f\"ur Maschinelle Sprachverarbeitung \\
  University of Stuttgart, Germany \\
  {\tt \{firstname.lastname\}@ims.uni-stuttgart.de}\\
}

\date{}

\begin{document}
\maketitle
\begin{abstract}
  The automatic detection of satire vs.\ regular news is relevant for
  downstream applications (for instance,
  knowledge base population) and to improve the understanding of
  linguistic characteristics of satire.  Recent approaches build upon
  corpora which have been labeled automatically based on article
  sources. We hypothesize that this encourages the models to learn
  characteristics for different publication sources (\eg, ``The Onion''
  vs.\ ``The Guardian'') rather than characteristics of satire,
  leading to poor generalization performance to unseen publication
  sources.  We therefore propose a novel model for satire detection
  with an adversarial component to control for the confounding
  variable of publication source.  On a large novel data set collected
  from German news (which we make available to the research
  community), we observe comparable satire classification performance
  and, as desired, a considerable drop in publication classification
  performance with adversarial training.  Our analysis shows that the
  adversarial component is crucial for the model to learn to pay
  attention to linguistic properties of satire.
\end{abstract}
\section{Introduction}
Satire is a form of art used to point out wrongs and criticize in an
entertaining manner \cite[\cf][p.\ 995ff.]{Sulzer1771}. It makes use
of different stylistic devices, \eg, humor, irony, sarcasm, 
exaggerations, parody or caricature
\citep{Knoche1982,Colletta2009}.  The occurrence of harsh and offensive as well as
banal and funny words is also typical
\citep{Golbert1962, Brummack1971}.

Satirical news are written with the aim of mimicking regular news in
diction. In contrast to misinformation and disinformation
\cite{Thorne2018}, it does not
have the intention of fooling the readers into actually believing
something wrong in order to manipulate their opinion.

The task of satire detection is to automatically distinguish satirical
news from regular news. This is relevant, for instance, for downstream
applications, such that satirical articles can be ignored in
knowledge base population. Solving this problem computationally is
challenging. Even human readers are sometimes not able to precisely
recognize satire \citep{Allcott2017}. Thus, an automatic system for
satire detection is both relevant for downstream applications and
could help humans to better understand the characteristics of satire.

Previous work mostly builds on top of corpora of news articles which
have been labeled automatically based on the publication source (\eg,
``The New York Times'' articles would be labeled as regular while
``The Onion'' articles as
satire\footnote{{https://www.theonion.com/},
  {https://www.nytimes.com/}}). We hypothesize that such distant
labeling approach leads to the model mostly representing
characteristics of the publishers instead of actual satire. This has
two main issues: First, interpretation of the model to obtain a better
understanding of concepts of satire would be misleading, and second,
generalization of the model to unseen publication sources would be
harmed.  We propose a new model with adversarial training to control
for the confounding variable of publication sources, \ie, we debias
the model.

Our experiments and analysis show that (1)~the satire detection
performance stays comparable when the adversarial component is
included, and (2) that adversarial training is crucial for the model
to pay attention to satire instead of publication characteristics.  In
addition, (3), we publish a large German data set for satire detection
which is a) the first data set in German, b) the first data set
including publication sources, enabling the experiments at hand, and
c) the largest resource for satire detection so
far.\footnote{Data/code:
  \hbox{\href{http://www.ims.uni-stuttgart.de/data/germansatire}{www.ims.uni-stuttgart.de/data/germansatire}}.}

\section{Previous Work}
Previous work tackled the task of automatic English satire detection
with handcrafted features, for instance, the validity of the context of
entity mentions \cite{Burfoot2009}, or the coherence of a story
\cite{Goldwasser2016}. \newcite{Rubin2016} use distributions of
parts-of-speech, sentiment, and exaggerations.  In contrast to these
approaches, our model uses only word embeddings as input
representations.
Our work is therefore similar to \newcite{Yang2017} and
\newcite{DeSarkar2018} who also use artificial neural networks to
predict if a given text is satirical or regular news. They develop a
hierarchical model of convolutional and recurrent layers with
attention over paragraphs or sentences. We follow this line of work but our model is not
hierarchical and introduces less parameters. We apply attention to
words instead of sentences or paragraphs, accounting for the fact that
satire might be expressed on a sub-sentence level.

Adversarial training is popular to improve the robustness of
models. Originally introduced by \newcite{Goodfellow2014} as
generative adversarial networks with a generative and a discriminative
component, \newcite{Ganin2016} show that a related concept can also be
used for domain adaptation: A domain-adversarial neural network
consists of a classifier for the actual class labels and a domain
discriminator. The two components share the same feature extractor and
are trained in a minimax optimization algorithm with gradient
reversal: The sign of the gradient of the domain discriminator is
flipped when backpropagating to the feature extractor. Building upon
the idea of eliminating domain-specific input representations,
\newcite{Wadsworth2018} debias input representations for recidivism
prediction, or income prediction
\cite{Edwards2016,Beutel2017,Madras2018,Zhang2018}.

Debiasing mainly focuses on word embeddings, \eg, to remove gender
bias from embeddings \cite{Bolukbasi2016}.  Despite previous positive
results with adversarial training, a recent study by
\citet{Elazar2018} calls for being cautious and not blindly trusting
adversarial training for debiasing.  We therefore analyze whether it
is possible at all to use adversarial training in another setting,
namely to control for the confounding variable of publication sources
in satire detection (see Section \ref{sec:hypothesis}).

\section{Methods for Satire Classification}
\subsection{Limitations of Previous Methods}
\label{sec:hypothesis}
The data set used by \citet{Yang2017} and \citet{DeSarkar2018}
consists of text from 14 satirical and 6 regular news
websites. Although the satire sources in train, validation, and test
sets did not overlap, the sources of regular news were not split up
according to the different data sets \cite{Yang2017}.  We
hypothesize that this enables the classifier to learn which articles
belong to which publication of regular news and classify everything else
as satire, given that one of the most frequent words is
the name of the website itself (see Section~\ref{sec:data}).
Unfortunately, we cannot analyze this potential limitation since their
data set does not contain any information on the publication
source\footnote{\url{https://data.mendeley.com/datasets/hx3rzw5dwt/draft?a=377d5571-af17-4e61-bf77-1b77b88316de},
  v.1, 2017, accessed on 2018-11-23}. Therefore, we create a new
corpus in German (see Section~\ref{sec:data}) including this
information and investigate our hypothesis on it.

\subsection{Model}
\label{sec:model}
Motivated by our hypothesis in Section \ref{sec:hypothesis}, we propose to consider two
different classification problems (satire detection and publication identification)
with a shared feature extractor. Figure \ref{fig:arch} provides an overview of our model.
We propose to train the publication identifier as an adversary.

\subsubsection{Feature Extractor}
Following \citet{DeSarkar2018}, we only use word embeddings and no
further handcrafted features to represent the input. We pretrain word
embeddings of 300 dimensions on the whole corpus using word2vec
\citep{Mikolov2013}. The feature generator $f$ takes the embeddings
of the words of each article as input
for a bidirectional LSTM \citep{Hochreiter1997}, followed by a
self-attention layer as proposed by \newcite{selfattention}.  We refer to the union of all the parameters of the
feature extractor as $\theta_f$ in the following.

\subsubsection{Satire Detector}
The gray part of Figure \ref{fig:arch} shows the model part for our
main task -- satire detection. 
The satire detector feeds the representation from the feature 
extractor into a softmax layer and performs a binary
classification task (satire: yes or no).
Note that, in contrast to \citet{DeSarkar2018}, we classify satire solely on the
document level, as this is sufficient to analyze the impact of the
adversarial component and the influence of the publication source.

\subsubsection{Publication Identifier}
The second classification branch of our model aims at identifying the publication source
of the input. Similar to the satire detector, the publication identifier consists
of a single softmax layer which gets the extracted features as an input.
It then performs a multi-class classification task since our dataset consists of 15 publication sources
(see Table \ref{tab:corpusstats}).

\subsubsection{Adversarial Training}
\begin{figure}
  \centering
  \includegraphics[width=.97\linewidth]{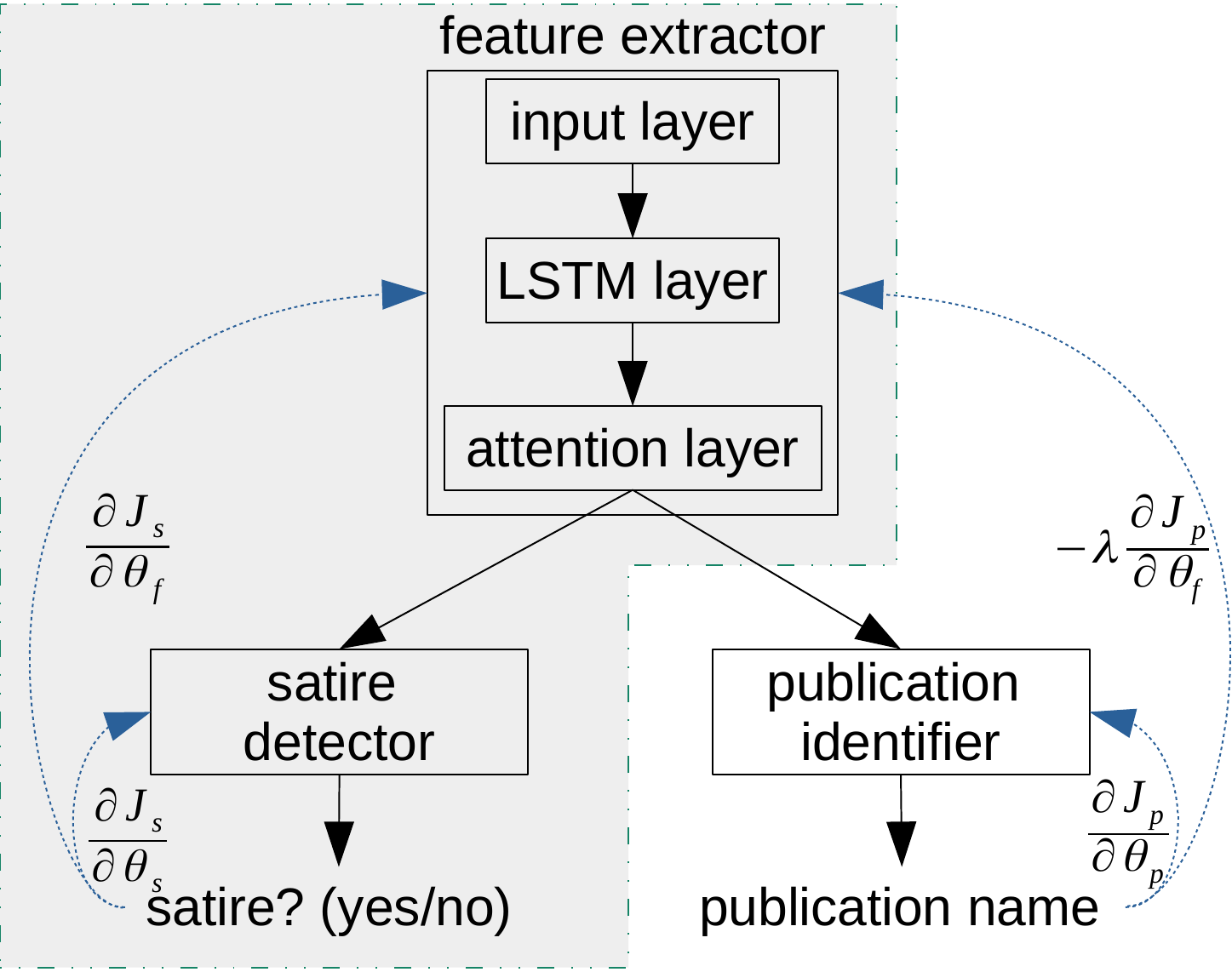}
  \caption{Architecture of the model. The gray area on the left
    shows the satire detector; the
    white area on the right is the adversary (publication identifier);
    the gradient flow with and without adversarial training is shown
    with blue arrows pointing upwards.}
  \label{fig:arch}
\end{figure}

Let $\theta_f$ be the parameters of the feature extractors
and $\theta_s$ and $\theta_p$ be the parameters of the satire
detector and the publication identifier, respectively.
The objective function for satire detection is
\begin{equation}
 J_s = - \mathbb{E}_{(x,y_s)\sim p_{\textrm{data}}}\log P_{\theta_f \cup \theta_s}(y_s,x)\,,
\end{equation}
while the objective for publication identification is
\begin{equation}
 J_p = - \mathbb{E}_{(x,y_p)\sim p_{\textrm{data}}}\log P_{\theta_f \cup \theta_p}(y_p,x)\,.
\end{equation}
Note that the parameters of the feature extractor $\theta_f$ are part of both model parts.
Since our goal is to control for the confounding variable of publication sources,
we train the publication identifier as an adversary: The parameters of the classification part $\theta_p$
are updated to optimize the publication identification while the parameters
of the shared feature generator $\theta_f$ are updated to fool the publication identifier.
This leads to the following update equations for the parameters
\begin{eqnarray}
\theta_s := \theta_s - \eta \frac{\partial J_s}{\partial \theta_s}\\
 \theta_p := \theta_p - \eta \frac{\partial J_p}{\partial \theta_p}\\
\theta_f := \theta_f - \eta \big(\frac{\partial J_s}{\partial \theta_f} -\lambda \frac{\partial J_p}{\partial \theta_f} \big)
\end{eqnarray}
with $\eta$ being the learning rate and  $\lambda$ being a weight for the 
reversed gradient that is tuned on the development set.
Figure \ref{fig:arch} depicts the gradient flow.

\section{Experiments}
\subsection{Experimental Setting}
\label{sec:data}
\textbf{Dataset.}
We consider German regular news collected from 4 websites and German satirical news
from 11 websites. Table~\ref{tab:corpusstats} shows statistics and
sources of the corpus, consisting of 
almost 330k articles.
The corpus contains articles published between January
1st, 2000 and May 1st, 2018. Each publication has
individual typical phrases and different most common words. 
Among the
most common words is typically the name of each publication, \eg, ``Der
Spiegel'' has ``SPIEGEL'' as fifth and ``Der Postillon'' ``Postillon'' as
third most common word. We did not delete those words to keep the dataset
as realistic as possible.
We randomly split the data set into training, development (dev) and
test (80/10/10 \%) with the same label distributions in all sets.
Given the comparable large size of the corpus, we
opt for using a well-defined test set for reproducability of our
experiments in contrast to a crossvalidation setting.

\begin{table}[t]
	\tabcolsep.8mm
        \footnotesize
        \centering
	\begin{tabular}{c|lrrrr}
	\toprule
	\multicolumn{1}{c}{}& & & \multicolumn{3}{c}{Average Length} \\
\cmidrule(ll){4-6}
	\multicolumn{1}{c}{}& Publication & \#Articles & Article & Sent. & Title \\
	\cmidrule(r){1-2}\cmidrule(l){3-3}\cmidrule(l){4-4}\cmidrule(l){5-5}\cmidrule(l){6-6}
	\multirow{4}{*}{\rotatebox[origin=c]{90}{Regular}} 
        & Der Spiegel & 31,180 & 556.74 & 19.04 & 7.47 \\
	& Der Standard & 53,632 & 328.82 & 18.62 & 6.3 \\
	& S\"udd. Zeit. & 177,605 & 635.77 & 17.58 & 7.74 \\
	& Die Zeit & 57,802 & 1,116.53 & 17.0 & 5.2 \\
        \cmidrule(r){1-2}\cmidrule(l){3-3}\cmidrule(l){4-4}\cmidrule(l){5-5}\cmidrule(l){6-6}
	\multirow{11}{*}{\rotatebox[origin=c]{90}{Satire}} 
        & Der Enth\"uller & 324 & 404.3 & 13.87 & 9.67 \\
	& Eulenspiegel & 192 & 1,072.17 & 17.45 & 4.38 \\
	& Nordd. Nach. & 211 & 188.46 & 17.84 & 8.46 \\
	& Der Postillon & 5,065 & 225.36 & 19.59 & 9.16 \\
	& Satirepatzer & 193 & 262.99 & 12.26 & 7.53 \\
	& Die Tagespresse & 1,271 & 301.28 & 16.39 & 10.83 \\
	& Titanic & 149 & 292.88 & 16.04 & 7.79 \\
	& Welt (Satire) & 1,249 & 291.45 & 21.76 & 9.02 \\
	& Der Zeitspiegel & 171 & 315.76 & 18.69 & 9.71 \\
	& Eine Zeitung & 416 & 265.16 & 16.04 & 13.35 \\
	& Zynismus24 & 402 & 181.59 & 17.67 & 11.96 \\
        \cmidrule(r){1-2}\cmidrule(l){3-3}\cmidrule(l){4-4}\cmidrule(l){5-5}\cmidrule(l){6-6}
	\multicolumn{1}{c}{}&\multicolumn{1}{l}{Regular} & 320,219 & 663.45 & 17.79 & 6.86 \\
	\multicolumn{1}{c}{}&\multicolumn{1}{l}{Satire} & 9,643 & 269.28 & 18.73 & 9.52 \\
	\bottomrule
	\end{tabular}
	\caption{Corpus statistics (average length in words)}
	\label{tab:corpusstats}
\end{table}

\textbf{Research questions.}
We discuss two questions.
RQ1: How does a decrease in publication classification performance
through adversarial training affect the satire classification performance?
RQ2: Is adversarial training effective for avoiding that the
model pays most attention to the characteristics of publication
source rather than actual satire?

\textbf{Baseline.} As a baseline model, we train the satire detector
part (gray area in Figure \ref{fig:arch}) on the satire task. Then, we
freeze the weights of the feature extractor and train the publication
classifier on top of it.  In addition, we use a majority baseline
model which predicts the most common class.

\textbf{Hyperparameters.}  We cut the input sentences to a maximum
length of 500 words. This enables us to fully represent almost all
satire articles and capture most of the content of the regular
articles while keeping the training time low. As mentioned before, we
represent the input words with 300 dimensional embeddings. The feature
extractor consists of a biLSTM layer with 300 hidden units in each
direction and a self-attention layer with an internal hidden
representation of 600. For training, we use Adam \cite{adam} with an
initial learning rate of 0.0001 and a decay rate of $10^{-6}$. We use
mini-batch gradient descent training with a batch size of 32 and
alternating batches of the two branches of our model. We avoid
overfitting by early stopping based on the satire F1 score on the
development set.

\textbf{Evaluation.} For evaluating satire detection, we use
precision, recall and F1 score of the satire class. For publication
identification, we calculate a weighted macro precision, recall and F1
score, \ie, a weighted sum of class-specific scores with weights
determined by the class distribution.

\subsection{Selection of Hyperparameter $\lambda$}
Table~\ref{tab:results} (upper part) shows results for different
values of $\lambda$, the hyperparameter of adversarial training, on
dev.  For $\lambda \in \left\{0.2, 0.3, 0.5\right\}$, the results are
comparably, with $\lambda=0.2$ performing best for satire
detection. Setting $\lambda=0.7$ leads to a performance drop for
satire but also to \F $=0$ for publication classification. Hence, we
chose $\lambda=0.2$ (the best performing model on satire
classification) and $\lambda=0.7$ (the worst performing model on
publication identification) to investigate RQ1.

\begin{table}[t]
  \tabcolsep1.0mm
  \centering
  \footnotesize
  \begin{tabular}{llcccccc}
    \toprule 
    && \multicolumn{3}{c}{Satire} & \multicolumn{3}{c}{Publication} \\
    \cmidrule(r){1-2}\cmidrule(lr){3-5}\cmidrule(l){6-8}
    & Model & P & R & F1 & P & R & F1 \\
    \cmidrule(r){1-2}\cmidrule(lr){3-5}\cmidrule(l){6-8}
    \multirow{6}{*}{\rotatebox{90}{dev}}
    &majority class & \pnull0.0 & \pnull0.0 & \pnull0.0 & 29.5 & 54.3 & 38.3 \\
    &no adv & 98.9 & \textbf{52.6} & \textbf{68.7} & 44.6 & 56.2 & 49.7 \\
    &adv, $\lambda=0.2$ & \textbf{99.3} & 50.8 & 67.2 & 31.2 & 55.4 & 40.0 \\
    &adv, $\lambda=0.3$ & 97.3 & 48.9 & 65.0 & 31.1 & 54.8 & 39.6 \\
    &adv, $\lambda=0.5$ & 99.1 & 50.8 & 67.2 & 31.7 & 55.2 & 40.3 \\
    &adv, $\lambda=0.7$ & 86.7 & 44.1 & 58.4 & \textbf{26.9} & \textbf{\pnull0.0} & \textbf{\pnull0.0} \\
    \cmidrule(r){1-2}\cmidrule(lr){3-5}\cmidrule(l){6-8}
    \multirow{4}{*}{\rotatebox{90}{test}}
    &majority class & \pnull0.0 & \pnull0.0 &\pnull0.0 & 29.1 & 53.9 & 37.8 \\
    &no adv. & 99.0 & \textbf{50.1} & \textbf{66.5} & 44.2 & 55.7 & 49.3 \\
    &adv, $\lambda=0.2$ & \textbf{99.4} & 49.4 & 66.0 & \textbf{30.8} & 54.8 & 39.5 \\
    &adv, $\lambda=0.7$ & 85.0 & 42.5 & 56.6 & 31.3 & \textbf{\pnull0.0} & \textbf{\pnull0.0} \\
    \bottomrule
  \end{tabular}
  \caption{Results on dev and independent test data.}
  \label{tab:results}
\end{table}

\section{Results (RQ1)}
\label{sec:resultsTest}
The bottom part of Table~\ref{tab:results} shows the results on test
data.  The majority baseline fails since the corpus contains more
regular than satirical news articles.  In comparison to the baseline
model without adversarial training (no adv), the model with
$\lambda=0.2$ achieves a comparable satire classification
performance. As expected, the publication identification performance
drops, especially the precision declines from $44.2$ \% to $30.8$
\%. Thus, a model which is punished for identifying publication
sources can still learn to identify satire.

Similar to the results on dev, the recall of the model with
$\lambda=0.7$ drops to (nearly) $0$ \%.  In this case, the satire
classification performance also drops.  This suggests that there are
overlapping features (cues) for both satire and publication
classification. This indicates that the two tasks cannot be entirely
untangled.

\begin{figure}[t]
  \newcommand{\noadv}{\rotatebox{90}{no adv}}
  \newcommand{\adv}{\rotatebox{90}{adv}}
  \newcommand{\tokenfont}{\sffamily\footnotesize\scalefont{0.9}}
  \emph{Example 1}\\ German original:\\
  \noadv\ \raisebox{5pt}{\fbox{\parbox{0.9\linewidth}{\tokenfont%
        \tw{Erfurt}{0} \tw{(}{0} \tw{dpo}{100} \tw{)}{0} \tw{-}{0}
        \tw{Es}{0} \tw{ist}{0} \tw{eine}{0} \tw{Organisation}{0}
        \tw{,}{0} \tw{die}{0} \tw{ausserhalb}{0} \tw{von}{0}
        \tw{Recht}{0} \tw{und}{0} \tw{Ordnung}{0} \tw{agiert}{0}
        \tw{,}{0} \tw{zahlreiche}{0} \tw{NPD-Funktion\"are}{0}
        \tw{finanziert}{0} \tw{und}{0} \tw{in}{0} \tw{nicht}{0}
        \tw{unerheblichem}{0} \tw{Ma\ss{}e}{0} \tw{in}{0} \tw{die}{0}
        \tw{Mordserie}{0} \tw{der}{0} \tw{sogenannten}{0}
        \tw{Zwickauer}{0} \tw{Zelle}{0} \tw{verstrickt}{0} \tw{ist}{0} \tw{.}{0}}}} \\
   \adv\ \raisebox{5pt}{\fbox{\parbox{0.9\linewidth}{\tokenfont%
        \tw{Erfurt}{0} \tw{(}{0} \tw{dpo}{0} \tw{)}{0} \tw{-}{0}
        \tw{Es}{0} \tw{ist}{0} \tw{eine}{0} \tw{Organisation}{0}
        \tw{,}{0} \tw{die}{0} \tw{ausserhalb}{0} \tw{von}{0}
        \tw{Recht}{0} \tw{und}{0} \tw{Ordnung}{0} \tw{agiert}{0}
        \tw{,}{0} \tw{zahlreiche}{0} \tw{NPD-Funktion\"are}{0}
        \tw{finanziert}{0} \tw{und}{0} \tw{in}{0} \tw{nicht}{1.4}
        \tw{unerheblichem}{7.8} \tw{Ma\ss{}e}{13.1} \tw{in}{15.7} \tw{die}{17.3}
        \tw{Mordserie}{14.1} \tw{der}{11.3} \tw{sogenannten}{8.2}
        \tw{Zwickauer}{5.2} \tw{Zelle}{3.1} \tw{verstrickt}{1.6}
        \tw{ist}{0.7} \tw{.}{0.5}}}} \\[1pt]
  English translation:\\
  \noadv\ \raisebox{5pt}{\fbox{\parbox{0.9\linewidth}{\tokenfont%
        \tw{Erfurt}{0} \tw{(}{0} \tw{dpo}{100} \tw{)}{0} \tw{-}{0}
        \tw{It}{0} \tw{is}{0} \tw{an}{0} \tw{organization}{0}
        \tw{which}{0} \tw{operates}{0} \tw{outside}{0} \tw{of}{0}
        \tw{law}{0} \tw{and}{0} \tw{order}{0} \tw{,}{0}
        \tw{funds}{0} \tw{numerous}{0} \tw{NPD}{0}
        \tw{operatives}{0} \tw{and}{0} \tw{is}{0} 
        \tw{to}{0} \tw{a}{0} \tw{not}{0} \tw{inconsiderable}{0} \tw{extent}{0}
        \tw{involved}{0}
        \tw{in}{0} \tw{the}{0} \tw{series}{0} \tw{of}{0}
        \tw{murders}{0} \tw{of}{0} \tw{the}{0} \tw{so}{0} \tw{called}{0}
        \tw{Zwickauer}{0} \tw{Zelle}{0}  \tw{.}{0}}}} \\
   \adv\ \raisebox{5pt}{\fbox{\parbox{0.9\linewidth}{\tokenfont%
        \tw{Erfurt}{0} \tw{(}{0} \tw{dpo}{0} \tw{)}{0} \tw{-}{0}
        \tw{It}{0} \tw{is}{0} \tw{an}{0} \tw{organization}{0}
        \tw{which}{0} \tw{operates}{0} \tw{outside}{0} \tw{of}{0}
        \tw{law}{0} \tw{and}{0} \tw{order}{0} \tw{,}{0}
        \tw{funds}{0} \tw{numerous}{0} \tw{NPD}{0}
        \tw{operatives}{0} \tw{and}{0} \tw{is}{0.7} \tw{to}{0} \tw{a}{0} \tw{not}{1.4} \tw{inconsiderable}{7.8} \tw{extent}{13.1} \tw{involved}{1.6}
        \tw{in}{15.7} \tw{the}{17.3} \tw{series}{14.1} \tw{of}{14.1}
        \tw{murders}{14.1} \tw{of}{11.3} \tw{the}{11.3} \tw{so}{8.2} \tw{called}{8.2}
        \tw{Zwickauer}{5.2} \tw{Zelle}{3.1} \tw{.}{0}}}} \\[2mm]
  \emph{Example 2}\\ German original:\\
	\noadv\ \raisebox{12pt}{\fbox{\parbox{0.9\linewidth}{\tokenfont%
        \tw{Immerhin}{4.2} \tw{wird}{4.2} \tw{derzeit}{4.2} \tw{der}{4.2} \tw{Vorschlag}{4.2}
        \tw{diskutiert}{4.2} \tw{,}{4.2} \tw{den}{4.2} \tw{Familiennachzug}{4.2}
        \tw{nur}{4.2} \tw{inklusive}{4.2} \tw{Schwiegerm\"uttern}{4.2} \tw{zu}{4.2}
        \tw{erlauben}{4.2} \tw{,}{4.2} \tw{wovon}{4.2} \tw{sich}{4.2}
        \tw{die}{4.2} \tw{Union}{4.2} \tw{einen}{4.2}
        \tw{abschreckenden}{4.2} \tw{Effekt}{4.2} \tw{erhofft}{4.2} \tw{.}{4.2}}}} \\
     \adv\ \raisebox{5pt}{\fbox{\parbox{0.9\linewidth}{\tokenfont%
        \tw{Immerhin}{0.4} \tw{wird}{0.7} \tw{derzeit}{1.2} \tw{der}{1.7} \tw{Vorschlag}{2.6}
        \tw{diskutiert}{4.3} \tw{,}{7.6} \tw{den}{12.4} \tw{Familiennachzug}{17.3}
        \tw{nur}{19.7} \tw{inklusive}{17.4} \tw{Schwiegerm\"uttern}{10.2} \tw{zu}{3.5}
        \tw{erlauben}{0.7} \tw{,}{0.1} \tw{wovon}{0} \tw{sich}{0}
        \tw{die}{0} \tw{Union}{0} \tw{einen}{0}
        \tw{abschreckenden}{0} \tw{Effekt}{0} \tw{erhofft}{0}
        \tw{.}{0}}}} \\[1pt]
  English translation:\\
	\noadv\ \raisebox{12pt}{\fbox{\parbox{0.9\linewidth}{\tokenfont%
        \tw{After}{4.2} \tw{all}{4.2} \tw{,}{4.2} \tw{the}{4.2} \tw{proposal}{4.2} \tw{to}{4.2} \tw{allow}{4.2}
        \tw{family}{4.2} \tw{reunion}{4.2} \tw{only}{4.2} \tw{inclusive}{4.2}
        \tw{mothers-in-law}{4.2} \tw{is}{4.2} \tw{being}{4.2} \tw{discussed}{4.2}
        \tw{,}{4.2} \tw{whereof}{4.2} \tw{the}{4.2} \tw{Union}{4.2}
        \tw{hopes}{4.2} \tw{for}{4.2} \tw{an}{4.2}
        \tw{off-putting}{4.2} \tw{effect}{4.2} \tw{.}{4.2}}}} \\
     \adv\ \raisebox{5pt}{\fbox{\parbox{0.9\linewidth}{\tokenfont%
        \tw{After}{0.4} \tw{all}{0.4} \tw{,}{0.0} \tw{the}{1.7} \tw{proposal}{2.6} \tw{to}{3.5} \tw{allow}{0.7}
        \tw{family}{17.3} \tw{reunion}{17.3} \tw{only}{19.7} \tw{inclusive}{17.4}
        \tw{mothers-in-law}{10.2} \tw{is}{0.7} \tw{being}{0.7} \tw{discussed}{4.3}
        \tw{,}{0.1} \tw{whereof}{0.0} \tw{the}{0.0} \tw{Union}{0.0}
        \tw{hopes}{0.0} \tw{for}{0.0} \tw{an}{0.0}
        \tw{off-putting}{0.0} \tw{effect}{0.0} \tw{.}{0.0}}}} \\
  \caption{Attention weight examples for satirical articles, with and without
    adversary.
    }
  \label{fig:attention}
\end{figure}

\section{Analysis (RQ2)}
To address RQ2, we analyze the results
and attention weights of the baseline model
and our model with adversarial training.
\subsection{Shift in Publication Identification}
\label{sec:pubidentification}
The baseline model (no adv) mostly predicts the correct publication
for a given article (in $55.7$ \% of the cases).  The model with
$\lambda=0.2$ mainly (in $98.2$ \% of the cases) predicts the most
common publication in our corpus (``S\"uddeutsche Zeitung'').  The
model with $\lambda=0.7$ shifts the majority of predictions ($98.7$
\%) to a rare class (namely ``Eine Zeitung''), leading to its bad
performance.

\subsection{Interpretation of Attention Weights}
\label{sec:attention}
Figure \ref{fig:attention} exemplifies the attention weights for a
selection of satirical instances. In the first example the baseline
model (no adv) focuses on a single word (``dpo'' as a parody of the
German newswire ``dpa'') which is unique to the publication the
article was picked from (``Der Postillon''). In comparison the model
using adversarial training ($\lambda=0.2$) ignores this word
completely and pays attention to ``die Mordserie'' (``series of
murders'') instead. In the second example, there are no words unique
to a publication and the baseline spreads the attention evenly across
all words. In contrast, the model with adversarial training is able to
find cues for satire, being humor in this example (``family reunion
[for refugees] is only allowed including mothers-in-law'').

\section{Conclusion and Future Work}
We presented evidence that simple neural networks for satire detection
learn to recognize characteristics of publication sources rather than
satire and proposed a model that uses adversarial training to control
for this effect. Our results show a considerable reduction of
publication identification performance while the satire detection
remains on comparable levels. The adversarial component enables the
model to pay attention to linguistic characteristics of satire. 

Future work could investigate the effect of other
potential confounding variables in satire detection, such as the
distribution of time and region of the articles. Further, we propose
to perform more quantitative but also more qualitative analysis to
better understand the behaviour of the two classifier configurations
in comparison.

\paragraph*{Acknowledgments}
This work has been partially funded by the German Research Council
(DFG), project KL 2869/1-1.

\newpage

\onecolumn

\section{Supplementary Material: Examples}
\newcommand{\noadv}{no\ adv}
\newcommand{\adv}{adv}
\begingroup
\flushleft
Example sentences with attention weights, with and without
adversarial training. `\noadv' refers to training without an
adversarial component, `\adv' refers to training with adversarial
component. For each example, we also provide an English translation. 
The weights of the translated words were set according to
the weights of the German words. If a German word was translated into
multiple English words, each of them received the original weight.\\[\baselineskip]
\endgroup
\noindent
  \begin{tabular}{ll}
  & Example (in German):\\
  \noadv &  \fbox{\parbox{0.8\linewidth}{\small%
        \tw{Erfurt}{0} \tw{(}{0} \tw{dpo}{100} \tw{)}{0} \tw{-}{0}
        \tw{Es}{0} \tw{ist}{0} \tw{eine}{0} \tw{Organisation}{0}
        \tw{,}{0} \tw{die}{0} \tw{ausserhalb}{0} \tw{von}{0}
        \tw{Recht}{0} \tw{und}{0} \tw{Ordnung}{0} \tw{agiert}{0}
        \tw{,}{0} \tw{zahlreiche}{0} \tw{NPD-Funktion\"are}{0}
        \tw{finanziert}{0} \tw{und}{0} \tw{in}{0} \tw{nicht}{0}
        \tw{unerheblichem}{0} \tw{Ma\ss{}e}{0} \tw{in}{0} \tw{die}{0}
        \tw{Mordserie}{0} \tw{der}{0} \tw{sogenannten}{0}
        \tw{Zwickauer}{0} \tw{Zelle}{0} \tw{verstrickt}{0} \tw{ist}{0} \tw{.}{0}}} \\
   \adv & \fbox{\parbox{0.8\linewidth}{\small%
        \tw{Erfurt}{0} \tw{(}{0} \tw{dpo}{0} \tw{)}{0} \tw{-}{0}
        \tw{Es}{0} \tw{ist}{0} \tw{eine}{0} \tw{Organisation}{0}
        \tw{,}{0} \tw{die}{0} \tw{ausserhalb}{0} \tw{von}{0}
        \tw{Recht}{0} \tw{und}{0} \tw{Ordnung}{0} \tw{agiert}{0}
        \tw{,}{0} \tw{zahlreiche}{0} \tw{NPD-Funktion\"are}{0}
        \tw{finanziert}{0} \tw{und}{0} \tw{in}{0} \tw{nicht}{1.4}
        \tw{unerheblichem}{7.8} \tw{Ma\ss{}e}{13.1} \tw{in}{15.7} \tw{die}{17.3}
        \tw{Mordserie}{14.1} \tw{der}{11.3} \tw{sogenannten}{8.2}
        \tw{Zwickauer}{5.2} \tw{Zelle}{3.1} \tw{verstrickt}{1.6}
        \tw{ist}{0.7} \tw{.}{0.5}}} \\
    \\
    
    & Translation into English:\\
    	\noadv &  \fbox{\parbox{0.8\linewidth}{\small%
        \tw{Erfurt}{0} \tw{(}{0} \tw{dpo}{100} \tw{)}{0} \tw{-}{0}
        \tw{It}{0} \tw{is}{0} \tw{an}{0} \tw{organization}{0}
        \tw{which}{0} \tw{operates}{0} \tw{outside}{0} \tw{of}{0}
        \tw{law}{0} \tw{and}{0} \tw{order}{0} \tw{,}{0}
        \tw{funds}{0} \tw{numerous}{0} \tw{NPD}{0}
        \tw{operatives}{0} \tw{and}{0} \tw{is}{0} 
        \tw{to}{0} \tw{a}{0} \tw{not}{0} \tw{inconsiderable}{0} \tw{extent}{0}
        \tw{involved}{0}
        \tw{in}{0} \tw{the}{0} \tw{series}{0} \tw{of}{0}
        \tw{murders}{0} \tw{of}{0} \tw{the}{0} \tw{so}{0} \tw{called}{0}
        \tw{Zwickauer}{0} \tw{Zelle}{0}  \tw{.}{0}}} \\
     \adv &  \fbox{\parbox{0.8\linewidth}{\small%
        \tw{Erfurt}{0} \tw{(}{0} \tw{dpo}{0} \tw{)}{0} \tw{-}{0}
        \tw{It}{0} \tw{is}{0} \tw{an}{0} \tw{organization}{0}
        \tw{which}{0} \tw{operates}{0} \tw{outside}{0} \tw{of}{0}
        \tw{law}{0} \tw{and}{0} \tw{order}{0} \tw{,}{0}
        \tw{funds}{0} \tw{numerous}{0} \tw{NPD}{0}
        \tw{operatives}{0} \tw{and}{0} \tw{is}{0.7} \tw{to}{0} \tw{a}{0} \tw{not}{1.4} \tw{inconsiderable}{7.8} \tw{extent}{13.1} \tw{involved}{1.6}
        \tw{in}{15.7} \tw{the}{17.3} \tw{series}{14.1} \tw{of}{14.1}
        \tw{murders}{14.1} \tw{of}{11.3} \tw{the}{11.3} \tw{so}{8.2} \tw{called}{8.2}
        \tw{Zwickauer}{5.2} \tw{Zelle}{3.1} \tw{.}{0}}} \\
\end{tabular}
      \begin{tabular}{ll}
    \\
	& Example (in German):\\
	\noadv &  \fbox{\parbox{0.8\linewidth}{\small%
        \tw{Oberfl\"achlich}{2.0} \tw{habe}{2.0} \tw{der}{2.0} \tw{jetzt}{2.0} \tw{als}{2.0}
        \tw{qualifiziert}{2.0} \tw{enttarnte}{2.0} \tw{Mann}{2.0} \tw{regelm\"a\ss{}ig}{2.0}
        \tw{Verdachtsmomente}{2.0} \tw{ignoriert}{2.0} \tw{und}{2.0} \tw{mit}{2.0}
        \tw{gro\ss{}er}{2.0} \tw{Sorgfalt}{2.0} \tw{Akten}{2.0} \tw{verloren}{2.0}
        \tw{oder}{2.0} \tw{versehentlich}{2.0} \tw{geschreddert}{2.0}
        \tw{:}{2.0} \tw{Erst}{2.0} \tw{nach}{2.0} \tw{seiner}{2.0}
        \tw{Enttarnung}{2.0} \tw{fand}{2.0} \tw{man}{2.0} \tw{heraus}{2.0} \tw{,}{2.0}
        \tw{dass}{2.0} \tw{der}{2.0} \tw{bei}{2.0} \tw{allen}{2.0} \tw{gesch\"atzte}{2.0}
        \tw{Kollege}{2.0} \tw{die}{2.0} \tw{vermeintlich}{2.0} \tw{verschlampten}{2.0} \tw{Akten}{2.0}
        \tw{oftmals}{2.0} \tw{heimlich}{2.0} \tw{wieder}{2.0} \tw{einsortierte}{2.0} \tw{und}{2.0}
        \tw{in}{2.0} \tw{unbeobachteten}{2.0} \tw{Momenten}{2.0} \tw{F\"alle}{2.0} \tw{bearbeitete}{2.0} \tw{.}{2.0}}} \\
	\adv &  \fbox{\parbox{0.8\linewidth}{\small%
        \tw{Oberfl\"achlich}{1.7} \tw{habe}{0.3} \tw{der}{0.1} \tw{jetzt}{0.0} \tw{als}{0.0}
        \tw{qualifiziert}{0.0} \tw{enttarnte}{0.0} \tw{Mann}{0.0} \tw{regelm\"a\ss{}ig}{0.0}
        \tw{Verdachtsmomente}{0.0} \tw{ignoriert}{0.0} \tw{und}{0} \tw{mit}{0}
        \tw{gro\ss{}er}{0.1} \tw{Sorgfalt}{0.2} \tw{Akten}{0.4} \tw{verloren}{0.7}
        \tw{oder}{1.0} \tw{versehentlich}{1.5} \tw{geschreddert}{2.5}
        \tw{:}{4.3} \tw{Erst}{7.2} \tw{nach}{10.2} \tw{seiner}{11.9}
        \tw{Enttarnung}{10.8} \tw{fand}{6.6} \tw{man}{2.4} \tw{heraus}{0.5} \tw{,}{0.1}
        \tw{dass}{0} \tw{der}{0} \tw{bei}{0} \tw{allen}{0} \tw{gesch\"atzte}{0}
        \tw{Kollege}{0} \tw{die}{0} \tw{vermeintlich}{0} \tw{verschlampten}{0} \tw{Akten}{0}
        \tw{oftmals}{0.1} \tw{heimlich}{0.2} \tw{wieder}{0.4} \tw{einsortierte}{0.6} \tw{und}{0.9}
        \tw{in}{1.4} \tw{unbeobachteten}{2.2} \tw{Momenten}{3.8}
        \tw{F\"alle}{6.4} \tw{bearbeitete}{9.5} \tw{.}{11.7}}} \\
\end{tabular}
      \begin{tabular}{ll}

         \\
    & Translation into English:\\
    	\noadv &  \fbox{\parbox{0.8\linewidth}{\small%
        \tw{Superficially}{2.0} \tw{,}{2.0} \tw{the}{2.0} \tw{now}{2.0} \tw{as}{2.0}
        \tw{skilled}{2.0} \tw{unmasked}{2.0} \tw{man}{2.0} \tw{consistently}{2.0}
        \tw{ignored}{2.0} \tw{suspicions}{2.0} \tw{and}{2.0} \tw{lost}{2.0} \tw{or}{2.0} \tw{accidentally}{1.5} \tw{shredded}{2.0}
        \tw{documents}{2.0} \tw{with}{2.0} \tw{high}{2.0} \tw{diligence}{2.0}
        \tw{:}{2.0} \tw{Only}{2.0} \tw{after}{2.0}
        \tw{his}{2.0} \tw{unmasking}{2.0} \tw{people}{2.0} \tw{found}{2.0}
        \tw{out}{2.0} \tw{that}{2.0} \tw{the}{2.0} \tw{by}{2.0}
        \tw{everyone}{2.0} \tw{valued}{2.0} \tw{colleague}{2.0} \tw{often}{2.0}
        \tw{secretly}{2.0} \tw{sorted}{2.0} \tw{the}{2.0} \tw{allegedly}{2.0} \tw{lost}{2.0}
        \tw{documents}{2.0} \tw{and}{2.0} \tw{processed}{2.0} \tw{them}{2.0} \tw{in}{2.0}
        \tw{unwatched}{2.0} \tw{moments}{2.0} \tw{.}{2.0}}} \\
	\adv &  \fbox{\parbox{0.8\linewidth}{\small%
        \tw{Superficially}{1.7} \tw{,}{0.0} \tw{the}{0.1} \tw{now}{0.0} \tw{as}{0.0}
        \tw{skilled}{0.0} \tw{unmasked}{0.0} \tw{man}{0.0} \tw{consistently}{0.0}
        \tw{ignored}{0.0} \tw{suspicions}{0.0} \tw{and}{0.0} \tw{lost}{0.7}
        \tw{or}{1.0} \tw{accidentally}{1.5} \tw{shredded}{2.5}
        \tw{documents}{0.4} \tw{with}{0.0} \tw{high}{0.1} \tw{diligence}{0.2}
        \tw{:}{4.3} \tw{Only}{7.2} \tw{after}{10.2}
        \tw{his}{11.9} \tw{unmasking}{10.8} \tw{people}{2.4} \tw{found}{6.6}
        \tw{out}{0.5} \tw{that}{0.0} \tw{the}{0.0} \tw{by}{0.0}
        \tw{everyone}{0.0} \tw{valued}{0.0} \tw{colleague}{0.0} \tw{often}{0.0}
        \tw{secretly}{0.2} \tw{sorted}{0.6} \tw{the}{0.0} \tw{allegedly}{0.0} \tw{lost}{0.0}
        \tw{documents}{0.0} \tw{and}{0.9} \tw{processed}{9.5} \tw{them}{0.0} \tw{in}{1.4}
        \tw{unwatched}{2.2} \tw{moments}{3.8} \tw{.}{11.7}}} \\
\\    
\end{tabular}
      \begin{tabular}{ll}
\\
\\
	& Example (in German):\\
	\noadv &  \fbox{\parbox{0.8\linewidth}{\small%
        \tw{Immerhin}{4.2} \tw{wird}{4.2} \tw{derzeit}{4.2} \tw{der}{4.2} \tw{Vorschlag}{4.2}
        \tw{diskutiert}{4.2} \tw{,}{4.2} \tw{den}{4.2} \tw{Familiennachzug}{4.2}
        \tw{nur}{4.2} \tw{inklusive}{4.2} \tw{Schwiegerm\"uttern}{4.2} \tw{zu}{4.2}
        \tw{erlauben}{4.2} \tw{,}{4.2} \tw{wovon}{4.2} \tw{sich}{4.2}
        \tw{die}{4.2} \tw{Union}{4.2} \tw{einen}{4.2}
        \tw{abschreckenden}{4.2} \tw{Effekt}{4.2} \tw{erhofft}{4.2} \tw{.}{4.2}}} \\
     \adv &  \fbox{\parbox{0.8\linewidth}{\small%
        \tw{Immerhin}{0.4} \tw{wird}{0.7} \tw{derzeit}{1.2} \tw{der}{1.7} \tw{Vorschlag}{2.6}
        \tw{diskutiert}{4.3} \tw{,}{7.6} \tw{den}{12.4} \tw{Familiennachzug}{17.3}
        \tw{nur}{19.7} \tw{inklusive}{17.4} \tw{Schwiegerm\"uttern}{10.2} \tw{zu}{3.5}
        \tw{erlauben}{0.7} \tw{,}{0.1} \tw{wovon}{0} \tw{sich}{0}
        \tw{die}{0} \tw{Union}{0} \tw{einen}{0}
        \tw{abschreckenden}{0} \tw{Effekt}{0} \tw{erhofft}{0} \tw{.}{0}}} \\
\\         
\end{tabular}
      \begin{tabular}{ll}
    & Translation into English:\\
    \noadv &  \fbox{\parbox{0.8\linewidth}{\small%
        \tw{After}{4.2} \tw{all}{4.2} \tw{,}{4.2} \tw{the}{4.2} \tw{proposal}{4.2} \tw{to}{4.2} \tw{allow}{4.2}
        \tw{family}{4.2} \tw{reunion}{4.2} \tw{only}{4.2} \tw{inclusive}{4.2}
        \tw{mothers-in-law}{4.2} \tw{is}{4.2} \tw{being}{4.2} \tw{discussed}{4.2}
        \tw{,}{4.2} \tw{whereof}{4.2} \tw{the}{4.2} \tw{Union}{4.2}
        \tw{hopes}{4.2} \tw{for}{4.2} \tw{an}{4.2}
        \tw{off-putting}{4.2} \tw{effect}{4.2} \tw{.}{4.2}}} \\
     \adv &  \fbox{\parbox{0.8\linewidth}{\small%
        \tw{After}{0.4} \tw{all}{0.4} \tw{,}{0.0} \tw{the}{1.7} \tw{proposal}{2.6} \tw{to}{3.5} \tw{allow}{0.7}
        \tw{family}{17.3} \tw{reunion}{17.3} \tw{only}{19.7} \tw{inclusive}{17.4}
        \tw{mothers-in-law}{10.2} \tw{is}{0.7} \tw{being}{0.7} \tw{discussed}{4.3}
        \tw{,}{0.1} \tw{whereof}{0.0} \tw{the}{0.0} \tw{Union}{0.0}
        \tw{hopes}{0.0} \tw{for}{0.0} \tw{an}{0.0}
        \tw{off-putting}{0.0} \tw{effect}{0.0} \tw{.}{0.0}}} \\
    \end{tabular}
    \begin{tabular}{ll}
\\
\\
        & Example (in German):\\
	\noadv &  \fbox{\parbox{0.8\linewidth}{\small%
        \tw{Berlin}{33.5} \tw{(}{23.8} \tw{EZ}{16.5} \tw{)}{11.1} \tw{|}{7.1} \tw{1}{4.2} \tw{.}{2.2} \tw{November}{1.0} \tw{2016}{0.4} \tw{|}{0.1} \tw{Nach}{0.0} \tw{Jahren}{0.0} \tw{des}{0.0} \tw{Streits}{0.0} \tw{gibt}{0.0} \tw{es}{0.0} \tw{endlich}{0.0} \tw{eine}{0.0} \tw{Einigung}{0.0} \tw{zwischen}{0.0} \tw{der}{0.0} \tw{Gema}{0.0} \tw{und}{0.0} \tw{Youtube}{0.0} \tw{.}{0.0} \tw{Wie}{1.7} \tw{heute}{1.7} \tw{bekannt}{1.8} \tw{wurde}{1.9} \tw{,}{2.1} \tw{d\"urfen}{2.3} \tw{zuk\"unftig}{2.6} \tw{wieder}{2.9} \tw{s\"amtliche}{3.2} \tw{Musikvideos}{3.6} \tw{auch}{4.1} \tw{in}{4.6} \tw{Deutschland}{5.1} \tw{abgespielt}{5.7} \tw{werden}{6.5} \tw{sofern}{7.5} \tw{auf}{8.5} \tw{die}{9.0} \tw{Tonspur}{8.4} \tw{verzichtet}{7.0} \tw{wird}{5.4} \tw{.}{4.4}}} \\
        \adv &  \fbox{\parbox{0.8\linewidth}{\small%
        \tw{Berlin}{8.0} \tw{(}{7.4} \tw{EZ}{7.2} \tw{)}{7.1}
        \tw{|}{7.1} \tw{1}{7.0} \tw{.}{6.8} \tw{November}{6.5}
        \tw{2016}{6.1} \tw{|}{5.7} \tw{Nach}{5.4} \tw{Jahren}{5.2}
        \tw{des}{5.0} \tw{Streits}{4.8} \tw{gibt}{4.2} \tw{es}{3.2}
        \tw{endlich}{1.9} \tw{eine}{0.9} \tw{Einigung}{0.4}
        \tw{zwischen}{0.2} \tw{der}{0.1} \tw{Gema}{0.0} \tw{und}{0.0}
        \tw{Youtube}{0.0} \tw{.}{0.0} \tw{Wie}{7.4} \tw{heute}{0.8}
        \tw{bekannt}{0.1} \tw{wurde}{0.0} \tw{,}{0.0} \tw{d\"urfen}{0.0}
        \tw{zuk\"unftig}{0.0} \tw{wieder}{0.0} \tw{s\"amtliche}{0.0}
        \tw{Musikvideos}{0.0} \tw{auch}{0.0} \tw{in}{0.0}
        \tw{Deutschland}{0.1} \tw{abgespielt}{0.2} \tw{werden}{0.5}
        \tw{sofern}{1.1} \tw{auf}{2.0} \tw{die}{3.4} \tw{Tonspur}{6.2}
        \tw{verzichtet}{12.3} \tw{wird}{24.1} \tw{.}{41.6}}} \\
         \\
\end{tabular}
      \begin{tabular}{ll}
    & Translation into English:\\
    \noadv &  \fbox{\parbox{0.8\linewidth}{\small%
	\tw{Berlin}{33.5} \tw{(}{23.8} \tw{EZ}{16.5} \tw{)}{11.1} \tw{|}{7.1} \tw{1}{4.2} \tw{.}{2.2} \tw{November}{1.0} \tw{2016}{0.4} \tw{|}{0.1} \tw{After}{0.0} \tw{years}{0.0} \tw{of}{0.0} \tw{fighting}{0.0} \tw{there}{0.0} \tw{finally}{0.0} \tw{is}{0.0} \tw{a}{0.0} \tw{settlement}{0.0} \tw{between}{0.0} \tw{the}{0.0} \tw{Gema}{0.0} \tw{and}{0.0} \tw{Youtube}{0.0} \tw{.}{0.0} \tw{It}{1.7} \tw{became}{1.7} \tw{known}{1.8} \tw{today}{1.7} \tw{,}{2.1} \tw{that}{0.0} \tw{in}{2.6} \tw{future}{2.6} \tw{every}{3.2} \tw{music}{3.6} \tw{video}{3.6} \tw{is}{2.3} \tw{allowed}{2.3} \tw{to}{6.5} \tw{be}{6.5} \tw{played}{5.7} \tw{back}{5.7} \tw{in}{4.6} \tw{Germany}{5.1} \tw{again}{2.9} \tw{,}{0.0} \tw{as}{7.5} \tw{long}{7.5} \tw{as}{7.5} \tw{the}{9.0} \tw{audio}{8.4} \tw{is}{5.4} \tw{removed}{7.0} \tw{.}{4.4}}} \\
	\adv &  \fbox{\parbox{0.8\linewidth}{\small%
	\tw{Berlin}{8.0} \tw{(}{7.4} \tw{EZ}{7.2} \tw{)}{7.1} \tw{|}{7.1} \tw{1}{7.0} \tw{.}{6.8} \tw{November}{6.5} \tw{2016}{6.1} \tw{|}{5.7} \tw{After}{5.4} \tw{years}{5.2} \tw{of}{5.0} \tw{fighting}{4.8} \tw{there}{3.2} \tw{finally}{1.9} \tw{is}{4.2} \tw{a}{0.9} \tw{settlement}{0.4} \tw{between}{0.2} \tw{the}{0.1} \tw{Gema}{0.0} \tw{and}{0.0} \tw{Youtube}{0.0} \tw{.}{0.0} \tw{It}{7.4} \tw{became}{0.0} \tw{known}{0.1} \tw{today}{0.8} \tw{,}{0.0} \tw{that}{0.0} \tw{in}{0.0} \tw{future}{0.0} \tw{every}{0.0} \tw{music}{0.0} \tw{video}{0.0} \tw{is}{0.0} \tw{allowed}{0.0} \tw{to}{0.5} \tw{be}{0.5} \tw{played}{0.2} \tw{back}{0.2} \tw{in}{0.0} \tw{Germany}{0.1} \tw{again}{0.0} \tw{,}{0.0} \tw{as}{1.1} \tw{long}{1.1} \tw{as}{1.1} \tw{the}{3.4} \tw{audio}{6.2} \tw{is}{24.1} \tw{removed}{12.3} \tw{.}{41.6}}} \\
\end{tabular}
      \begin{tabular}{ll}
\\
\\
        & Example (in German):\\
	\noadv &  \fbox{\parbox{0.8\linewidth}{\small%
\tw{Stuttgart}{33.5} \tw{Die}{23.8} \tw{Polizeidirektion}{16.5} \tw{Stuttgart}{11.1} \tw{lobte}{7.1} \tw{die}{4.2} \tw{Beamten}{2.2} \tw{ausdr\"ucklich}{1.0} \tw{f\"ur}{0.4} \tw{ihren}{0.1} \tw{\"uberdurchschnittlichen}{0.0} \tw{Einsatz}{0.0} \tw{und}{0.0} \tw{die}{0.0} \tw{Bereitschaft}{0.0} \tw{,}{0.0} \tw{sogar}{0.0} \tw{auf}{0.0} \tw{einer}{0.0} \tw{Betriebsfeier}{0.0} \tw{noch}{0.0} \tw{weiter}{0.0} \tw{zu}{0.0} \tw{arbeiten}{0.0} \tw{.}{0.0} \tw{Kritische}{0.0} \tw{Stimmen}{0.0} \tw{wenden}{0.0} \tw{ein}{0.0} \tw{,}{0.0} \tw{dass}{0.0} \tw{die}{0.0} \tw{Beamten}{0.0} \tw{ganz}{0.0} \tw{offensichtlich}{0.0} \tw{eine}{0.0} \tw{wilde}{0.0} \tw{Party}{0.0} \tw{mit}{0.0} \tw{beschlagnahmten}{0.0} \tw{Drogen}{0.0} \tw{gefeiert}{0.3} \tw{h\"atten}{4.4} \tw{.}{95.2}}} \\
	\adv &  \fbox{\parbox{0.8\linewidth}{\small%
\tw{Stuttgart}{8.0} \tw{Die}{7.4} \tw{Polizeidirektion}{7.2} \tw{Stuttgart}{7.1} \tw{lobte}{7.1} \tw{die}{7.0} \tw{Beamten}{6.8} \tw{ausdr\"ucklich}{6.5} \tw{f\"ur}{6.1} \tw{ihren}{5.7} \tw{\"uberdurchschnittlichen}{5.4} \tw{Einsatz}{5.2} \tw{und}{5.0} \tw{die}{4.8} \tw{Bereitschaft}{4.2} \tw{,}{3.2} \tw{sogar}{1.9} \tw{auf}{0.9} \tw{einer}{0.4} \tw{Betriebsfeier}{0.2} \tw{noch}{0.1} \tw{weiter}{0.0} \tw{zu}{0.0} \tw{arbeiten}{0.0} \tw{.}{0.0} \tw{Kritische}{0.4} \tw{Stimmen}{0.1} \tw{wenden}{0.1} \tw{ein}{0.1} \tw{,}{0.1} \tw{dass}{0.1} \tw{die}{0.1} \tw{Beamten}{0.1} \tw{ganz}{0.2} \tw{offensichtlich}{0.2} \tw{eine}{0.5} \tw{wilde}{1.0} \tw{Party}{2.2} \tw{mit}{4.2} \tw{beschlagnahmten}{6.4} \tw{Drogen}{9.1} \tw{gefeiert}{13.6} \tw{h\"atten}{22.6} \tw{.}{38.7}}} \\	        
\end{tabular}
      \begin{tabular}{ll}
\\
    & Translation into English:\\
 \noadv &  \fbox{\parbox{0.8\linewidth}{\small%
\tw{Stuttgart}{33.5} \tw{The}{23.8} \tw{police}{16.5} \tw{administration}{16.5} \tw{in}{11.1} \tw{Stuttgart}{11.1} \tw{commended}{7.1} \tw{the}{4.2} \tw{policemen}{2.2} \tw{explicitly}{1.0} \tw{for}{0.4} \tw{their}{0.1} \tw{outstanding}{0.0} \tw{commitment}{0.0} \tw{and}{0.0} \tw{their}{0.0} \tw{willingness}{0.0} \tw{to}{0.0} \tw{continue}{0.0} \tw{working}{0.0} \tw{even}{0.0} \tw{on}{0.0} \tw{a}{0.0} \tw{work}{0.0} \tw{party}{0.0} \tw{.}{0.0} \tw{Critical}{0.0} \tw{voices}{0.0} \tw{argue}{0.0} \tw{that}{0.0} \tw{the}{0.0} \tw{policemen}{0.0} \tw{obviously}{0.0} \tw{had}{4.4} \tw{thrown}{4.4} \tw{a}{0.0} \tw{wild}{0.0} \tw{party}{0.0} \tw{with}{0.0} \tw{the}{0.0} \tw{seized}{0.0} \tw{drugs}{0.0} \tw{.}{95.2}}} \\
	\adv &  \fbox{\parbox{0.8\linewidth}{\small%
\tw{Stuttgart}{8.0} \tw{The}{7.4} \tw{police}{7.2} \tw{administration}{7.2} \tw{in}{7.1} \tw{Stuttgart}{7.1} \tw{commended}{7.1} \tw{the}{7.0} \tw{policemen}{6.8} \tw{explicitly}{6.5} \tw{for}{6.1} \tw{their}{5.7} \tw{outstanding}{5.4} \tw{commitment}{5.2} \tw{and}{5.0} \tw{their}{4.8} \tw{willingness}{4.2} \tw{to}{0.0} \tw{continue}{0.0} \tw{working}{0.0} \tw{even}{1.9} \tw{on}{0.9} \tw{a}{0.4} \tw{work}{0.2} \tw{party}{0.2} \tw{.}{0.0} \tw{Critical}{0.4} \tw{voices}{0.1} \tw{argue}{0.1} \tw{that}{0.1} \tw{the}{0.1} \tw{policemen}{0.1} \tw{obviously}{0.2} \tw{had}{22.6} \tw{thrown}{13.6} \tw{a}{0.5} \tw{wild}{1.0} \tw{party}{2.2} \tw{with}{4.2} \tw{the}{4.2} \tw{seized}{6.4} \tw{drugs}{9.1} \tw{.}{38.7}}} \\    

    \end{tabular}


\begin{thebibliography}{25}
\expandafter\ifx\csname natexlab\endcsname\relax\def\natexlab#1{#1}\fi

\bibitem[{Allcott and Gentzkow(2017)}]{Allcott2017}
Hunt Allcott and Matthew Gentzkow. 2017.
\newblock \href {http://doi.org/10.1257/jep.31.2.211} {{Social Media and Fake
  News in the 2016 Election}}.
\newblock \emph{Journal of Economic Perspectives}, 31(2):211--236.

\bibitem[{Beutel et~al.(2017)Beutel, Chen, Zhao, and Chi}]{Beutel2017}
Alex Beutel, Jilin Chen, Zhe Zhao, and Ed~H. Chi. 2017.
\newblock \href {https://arxiv.org/pdf/1707.00075.pdf} {{Data Decisions and
  Theoretical Implications when Adversarially Learning Fair Representations}}.
\newblock In \emph{4th Workshop on Fairness, Accountability, and Transparency
  in Machine Learning (FAT/ML) at 23rd SIGKDD conference on Knowledge Discovery
  and Data Mining (KDD 2017)}, Halifax, Canada.

\bibitem[{Bolukbasi et~al.(2016)Bolukbasi, Chang, Zou, Saligrama, and
  Kalai}]{Bolukbasi2016}
Tolga Bolukbasi, Kai-Wei Chang, James~Y Zou, Venkatesh Saligrama, and Adam~T
  Kalai. 2016.
\newblock \href
  {https://papers.nips.cc/paper/6228-man-is-to-computer-programmer-as-woman-is-to-homemaker-debiasing-word-embeddings.pdf}
  {{Man is to Computer Programmer as Woman is to Homemaker? Debiasing Word
  Embeddings}}.
\newblock In \emph{Advances in Neural Information Processing Systems}, pages
  4349--4357.

\bibitem[{Brummack(1971)}]{Brummack1971}
J\"urgen Brummack. 1971.
\newblock \href {https://doi.org/10.1007/BF03376186} {{Zu Begriff und Theorie
  der Satire}}.
\newblock \emph{Deutsche Vierteljahrsschrift f\"ur Literaturwissenschaft und
  Geistesgeschichte}, 45(1):275--377.
\newblock In German.

\bibitem[{Burfoot and Baldwin(2009)}]{Burfoot2009}
Clint Burfoot and Timothy Baldwin. 2009.
\newblock \href {http://aclweb.org/anthology/P09-2041} {{Automatic Satire
  Detection: Are You Having a Laugh?}}
\newblock In \emph{Proceedings of the ACL-IJCNLP 2009 Conference Short Papers},
  pages 161--164. Association for Computational Linguistics.

\bibitem[{Colletta(2009)}]{Colletta2009}
Lisa Colletta. 2009.
\newblock \href {https://doi.org/10.1111/j.1540-5931.2009.00711.x} {{Political
  Satire and Postmodern Irony in the Age of Stephen Colbert and Jon Stewart}}.
\newblock \emph{The Journal of Popular Culture}, 42(5):856--874.

\bibitem[{De~Sarkar et~al.(2018)De~Sarkar, Yang, and Mukherjee}]{DeSarkar2018}
Sohan De~Sarkar, Fan Yang, and Arjun Mukherjee. 2018.
\newblock \href {http://aclweb.org/anthology/C18-1285} {{Attending Sentences to
  detect Satirical Fake News}}.
\newblock In \emph{Proceedings of the 27th International Conference on
  Computational Linguistics}, pages 3371--3380. Association for Computational
  Linguistics.

\bibitem[{Edwards and Storkey(2016)}]{Edwards2016}
Harrison Edwards and Amos Storkey. 2016.
\newblock \href {http://arxiv.org/abs/1511.05897} {{Censoring Representations
  with an Adversary}}.
\newblock In \emph{International Conference on Learning Representations}.

\bibitem[{Elazar and Goldberg(2018)}]{Elazar2018}
Yanai Elazar and Yoav Goldberg. 2018.
\newblock \href {http://www.aclweb.org/anthology/D18-1002} {Adversarial removal
  of demographic attributes from text data}.
\newblock In \emph{Proceedings of the 2018 Conference on Empirical Methods in
  Natural Language Processing}, pages 11--21, Brussels, Belgium. Association
  for Computational Linguistics.

\bibitem[{Ganin et~al.(2016)Ganin, Ustinova, Ajakan, Germain, Larochelle,
  Laviolette, Marchand, and Lempitsky}]{Ganin2016}
Yaroslav Ganin, Evgeniya Ustinova, Hana Ajakan, Pascal Germain, Hugo
  Larochelle, Fran\c{c}ois Laviolette, Mario Marchand, and Victor Lempitsky.
  2016.
\newblock \href {http://dl.acm.org/citation.cfm?id=2946645.2946704}
  {{Domain-adversarial Training of Neural Networks}}.
\newblock \emph{Journal of Machine Learning Research}, 17(1):2030--2096.

\bibitem[{Golbert(1962)}]{Golbert1962}
Highet Golbert. 1962.
\newblock \emph{{The Anatomy of Satire}}.
\newblock Princeton paperbacks. Princeton University Press.

\bibitem[{Goldwasser and Zhang(2016)}]{Goldwasser2016}
Dan Goldwasser and Xiao Zhang. 2016.
\newblock \href {http://aclweb.org/anthology/Q16-1038} {{Understanding
  Satirical Articles Using Common-Sense}}.
\newblock \emph{Transactions of the Association for Computational Linguistics},
  4:537--549.

\bibitem[{Goodfellow et~al.(2014)Goodfellow, Pouget-Abadie, Mirza, Xu,
  Warde-Farley, Ozair, Courville, and Bengio}]{Goodfellow2014}
Ian~J. Goodfellow, Jean Pouget-Abadie, Mehdi Mirza, Bing Xu, David
  Warde-Farley, Sherjil Ozair, Aaron Courville, and Yoshua Bengio. 2014.
\newblock \href {http://dl.acm.org/citation.cfm?id=2969033.2969125}
  {{Generative Adversarial Nets}}.
\newblock In \emph{Proceedings of the 27th International Conference on Neural
  Information Processing Systems - Volume 2}, NIPS'14, pages 2672--2680,
  Cambridge, MA, USA. MIT Press.

\bibitem[{Hochreiter and Schmidhuber(1997)}]{Hochreiter1997}
Sepp Hochreiter and J\"urgen Schmidhuber. 1997.
\newblock \href {https://doi.org/10.1162/neco.1997.9.8.1735} {{Long Short-Term
  Memory}}.
\newblock \emph{Neural Computation}, 9(8):1735--1780.

\bibitem[{Kingma and Ba(2014)}]{adam}
Diederik~P. Kingma and Jimmy Ba. 2014.
\newblock \href {http://arxiv.org/abs/1412.6980} {Adam: A method for stochastic
  optimization}.
\newblock In \emph{International Conference on Learning Representations}.

\bibitem[{Knoche(1982)}]{Knoche1982}
Ulrich Knoche. 1982.
\newblock \emph{{Die r{\"o}mische Satire}}.
\newblock Orbis Biblicus Et Orientalis - Series Archaeologica. Vandenhoeck \&
  Ruprecht.
\newblock In German.

\bibitem[{Lin et~al.(2017)Lin, Feng, dos Santos, Yu, Xiang, Zhou, and
  Bengio}]{selfattention}
Zhouhan Lin, Minwei Feng, Cicero~Nogueira dos Santos, Mo~Yu, Bing Xiang, Bowen
  Zhou, and Yoshua Bengio. 2017.
\newblock \href {https://openreview.net/pdf?id=BJC_jUqxe} {A structured
  self-attentive sentence embedding}.
\newblock In \emph{International Conference on Learning Representations}.

\bibitem[{Madras et~al.(2018)Madras, Creager, Pitassi, and Zemel}]{Madras2018}
David Madras, Elliot Creager, Toniann Pitassi, and Richard Zemel. 2018.
\newblock \href {http://proceedings.mlr.press/v80/madras18a.html} {Learning
  adversarially fair and transferable representations}.
\newblock In \emph{Proceedings of the 35th International Conference on Machine
  Learning}, volume~80 of \emph{Proceedings of Machine Learning Research},
  pages 3384--3393, Stockholmsm{\"a}ssan, Stockholm Sweden. PMLR.

\bibitem[{Mikolov et~al.(2013)Mikolov, Chen, Corrado, and Dean}]{Mikolov2013}
Tomas Mikolov, Kai Chen, Greg Corrado, and Jeffrey Dean. 2013.
\newblock \href {https://arxiv.org/abs/1301.3781} {Efficient estimation of word
  representations in vector space}.
\newblock In \emph{Workshop at International Conference on Learning
  Representations}.

\bibitem[{Rubin et~al.(2016)Rubin, Conroy, Chen, and Cornwell}]{Rubin2016}
Victoria Rubin, Niall Conroy, Yimin Chen, and Sarah Cornwell. 2016.
\newblock \href {http://www.aclweb.org/anthology/W16-0802} {Fake news or truth?
  using satirical cues to detect potentially misleading news}.
\newblock In \emph{Proceedings of the Second Workshop on Computational
  Approaches to Deception Detection}, pages 7--17, San Diego, California.
  Association for Computational Linguistics.

\bibitem[{Sulzer(1771)}]{Sulzer1771}
Johann~Georg Sulzer. 1771.
\newblock \href
  {http://www.deutschestextarchiv.de/book/view/sulzer_theorie02_1774?p=424}
  {\emph{{Allgemeine Theorie der Sch\"onen K\"unste}}}, 1. edition.
\newblock Weidmann; Reich, Leipzig.
\newblock In German.

\bibitem[{Thorne and Vlachos(2018)}]{Thorne2018}
James Thorne and Andreas Vlachos. 2018.
\newblock \href {http://www.aclweb.org/anthology/C18-1283} {{Automated Fact
  Checking: Task Formulations, Methods and Future Directions}}.
\newblock In \emph{Proceedings of the 27th International Conference on
  Computational Linguistics}, pages 3346--3359, Santa Fe, New Mexico, USA.
  Association for Computational Linguistics.

\bibitem[{Wadsworth et~al.(2018)Wadsworth, Vera, and Piech}]{Wadsworth2018}
Christina Wadsworth, Francesca Vera, and Chris Piech. 2018.
\newblock \href
  {http://www.fatml.org/media/documents/achieving_fairness_through_adversearial_learning.pdf}
  {Achieving fairness through adversarial learning: an application to
  recidivism prediction}.
\newblock In \emph{Proceedings of the Conference on Fairness, Accountability,
  and Transparency in Machine Learning (FATML)}, Stockholm, Sweden.

\bibitem[{Yang et~al.(2017)Yang, Mukherjee, and Dragut}]{Yang2017}
Fan Yang, Arjun Mukherjee, and Eduard Dragut. 2017.
\newblock \href {https://doi.org/10.18653/v1/D17-1211} {{Satirical News
  Detection and Analysis using Attention Mechanism and Linguistic Features}}.
\newblock In \emph{Proceedings of the 2017 Conference on Empirical Methods in
  Natural Language Processing}, pages 1979--1989. Association for Computational
  Linguistics.

\bibitem[{Zhang et~al.(2018)Zhang, Lemoine, and Mitchell}]{Zhang2018}
Brian~Hu Zhang, Blake Lemoine, and Margaret Mitchell. 2018.
\newblock \href {https://doi.org/10.1145/3278721.3278779} {Mitigating unwanted
  biases with adversarial learning}.
\newblock In \emph{Proceedings of the 2018 AAAI/ACM Conference on AI, Ethics,
  and Society}, AIES '18, pages 335--340, New York, NY, USA. ACM.

\end{thebibliography}
\end{document}